%% file: main.tex
\documentclass[11pt,a4paper]{article}
\usepackage[final]{neurips_2018}

\usepackage[utf8]{inputenc} % allow utf-8 input
\usepackage[T1]{fontenc}    % use 8-bit T1 fonts
\usepackage{hyperref}       % hyperlinks
\usepackage{url}            % simple URL typesetting
\usepackage{booktabs}       % professional-quality tables
\usepackage{amsfonts}       % blackboard math symbols
\usepackage{nicefrac}       % compact symbols for 1/2, etc.
\usepackage{microtype}      % microtypography
\usepackage{graphicx}
\usepackage{etoolbox}
\usepackage{multirow}
\usepackage{tikz}
\usepackage{authblk}

\usepackage{amsmath}

\title{Towards leveraging LLMs for Conditional  QA}

\author{\textbf{Syed-Amad Hussain\textsuperscript{1}, Parag Pravin Dakle\textsuperscript{2}, SaiKrishna Rallabandi\textsuperscript{2} and Preethi Raghavan\textsuperscript{2}} \\
  \textsuperscript{1} Ohio State University, Columbus, OH \\
  \textsuperscript{2}AI Center of Excellence, Fidelity Investments, Boston MA\\
  \textit{hussain.97@buckeyemail.osu.edu} \\
  \{\texttt{paragpravin.dakle, saikrishna.rallabandi, preethi.raghavan\}@fmr.com}}

\begin{document}
\renewcommand{\arraystretch}{1.3}
\maketitle

\begin{abstract}
\input{sections/abstract.tex}
\end{abstract}

%%%%%%%%%%%% Introduction
\input{sections/introduction}
\input{sections/related_works}

%%% Methods
\input{sections/methods}

%%% Experimental Setup
\input{sections/experiments}

%%% Results
\input{sections/results}

%% Analysis
\input{sections/analysis}

%%% Conclusion
\input{sections/conclusion}

%%%%%%%%%%%% Bibliography
\bibliographystyle{plainnat}
\bibliography{refs}

\input{sections/appendix}

\end{document}

%% file: sections/abstract.tex
%While LLMs have excelled in various question-answering tasks, challenges persist in handling diverse question types and conversational contexts, including complex and ambiguous questions, memory bias, document grounding, and dealing with incomplete information. We examine the performance of Large Language Models (LLMs) in conditional question-answering tasks, with a focus on the CQA dataset and the use of generative LLMs like T5 and UL2. We find that surpassing the state-of-the-art (SOTA) is achievable in some cases through fine-tuned LLMs, even without fully encoding all input context. However, generative models struggle with extractive answers and with the potential to inject false information. A study with oracle-retrievers finds that performance is heavily dependent on effective retrieval, promoting the need for more advanced solutions. Additionally, we emphasize the influence of evaluation metrics on performance assessments, noting discrepancies in ranking models. We argue for a more comprehensive evaluation framework and highlight the difficulty of assessing performance without human baselines for comparison, indicating the ongoing complexity of the task. We propose future work seeks to manipulate the training task to prioritize evidence extraction before answer generation, encouraging better evidence retrieval and utilization. We also suggest the exploration of prompt-based techniques, such as self-correction and few-shot answer generation, to further enhance LLM performance in conditional question-answering tasks.

This study delves into the capabilities and limitations of Large Language Models (LLMs) in the challenging domain of conditional question-answering. Utilizing the Conditional Question Answering (CQA) dataset and focusing on generative models like T5 and UL2, we assess the performance of LLMs across diverse question types. Our findings reveal that fine-tuned LLMs can surpass the state-of-the-art (SOTA) performance in some cases, even without fully encoding all input context, with an increase of 7-8 points in Exact Match (EM) and F1 scores for Yes/No questions. However, these models encounter challenges in extractive question answering, where they lag behind the SOTA by over 10 points, and in mitigating the risk of injecting false information. A study with oracle-retrievers emphasizes the critical role of effective evidence retrieval, underscoring the necessity for advanced solutions in this area. Furthermore, we highlight the significant influence of evaluation metrics on performance assessments and advocate for a more comprehensive evaluation framework. The complexity of the task, the observed performance discrepancies, and the need for effective evidence retrieval underline the ongoing challenges in this field and underscore the need for future work focusing on refining training tasks and exploring prompt-based techniques to enhance LLM performance in conditional question-answering tasks.

%% file: sections/introduction.tex
\section{Introduction}
\label{introduction}

Contextual question answering (CQA) is a significant area of research in Natural Language Processing, with a multitude of challenges to be addressed for achieving high-performing systems. In CQA, questions are often based on a given context, and the responses should adhere to this context rather than relying solely on pre-existing knowledge. Large Language Models (LLMs) have demonstrated impressive performance across diverse tasks, including CQA, significantly revolutionizing the field of natural language processing.
Despite their notable performance, LLMs face several areas for improvement, especially in handling diverse types of questions and conversational contexts. Key challenges persist in understanding complex and ambiguous questions, dealing with memory bias and document grounding, and managing incomplete or missing information. Our findings reveal that fine-tuned LLMs can surpass state-of-the-art (SOTA) performance in some areas, with an increase of 7-8 points in Exact Match (EM) and F1 scores for Yes/No questions. However, the models encounter challenges in extractive question answering, lagging behind the SOTA by over 10 points.
Our investigation delves into the performance of LLMs across various question types, including extractive, generative, unanswerable, multi-hop, and incomplete questions. We also explore context retrieval mechanisms and the potential of fine-tuning strategies for evidence-based generation. Despite recent LLMs showing exceptional performance in many tasks, we posit that a fine-tuned approach can achieve state-of-the-art performance, generating accurate, complete responses with justification via extracted evidence.
Evaluation techniques employed in our study range from traditional metrics such as EM and F1 scores to embedding-based methods such as Cosine Similarity, Bert Score, and Bart Score. This diverse evaluation approach aims to offer a comprehensive understanding of the strengths and weaknesses of LLMs in the context of CQA.
In conclusion, this study presents an extensive exploration of LLMs in the task of CQA, highlighting their performance across various question types, context retrieval methods, fine-tuning strategies, and evaluation techniques. Despite significant challenges, our findings suggest potential avenues for enhancing the performance of LLMs in conditional question answering tasks, contributing to the ongoing efforts to improve the reliability and versatility of AI systems.

%% file: sections/related_works.tex
\section{Related Works}

\subsection{Question Answering Tasks}
Since the introduction of SQuAD, many question answering (QA) datasets have been proposed, contributing to significant improvements in QA model performance (\cite{SQuaD-Rajpurkar2016, Rajpurkar2018, Yang2018, naturalQuestions-kwiatkowski2019, Ferguson2020, Dasigi2021}). Additionally, the utilization of large pretrained language models has enhanced traditional reading comprehension and QA tasks (\cite{devlin2018BERT, t5, brown2020, rae2022scaling, zhang2022opt, chowdhery2022palm, chung2022scaling}). To introduce greater complexity, researchers have released several multi-hop QA datasets designed to evaluate models' capacity to solve intricate questions (\cite{Yang2018, Ferguson2020}). Further work seeks to expand this reasoning and contextualization over long-document content or, in the case of conversational QA, long conversational-history content (\cite{Dasigi2021, reddy2019coqa}. More recently, ambiguity has been introduced into QA datasets so language models are challenged to resolve ambiguous questions, whether that is due to phrasing issues, missing information, or inherent non-determinism in the query (\cite{ambigQA-Min2020, cqa}. In this work, we make use of the Conditional QA dataset which seeks to be comprehensive to each of these major QA dataset augmentations, while also being domain-specific, supporting challenges in semantic drift between open and closed-domain queries (\cite{cqa}).

\subsection{Document Grounded QA}
Effective QA over long input content requires the QA system to comprehend facts spread throughout the long context. Typically, this can be done either through a multi-stage retrieve-then-read paradigm or a single end-to-end model that digests the entire input context and generates a response (\cite{bohnet2023attributed}. Retrieve-then-read can be implemented using dual encoders, neural indexing, and most recently, few-shot prompting of an LLM (\cite{ChenFWB17,ni2021,chowdhery2022palm}. These methods allow for disparate levels of reasoning between evidence, as certain facts may be missed by the retriever in cases where the fact is only pertinent when reasoning over the query and another fact. Dense end-to-end approaches, that directly read the entire input context and generate a response, have found success in this task, despite having to make architectural changes to handle arbitrarily long inputs (\cite{izacard2021leveraging}. Similarly, dense models have also found success in closed-book QA tasks where there is no provided input context, challenging the model to use its latent memory gathered in pre-training and fine-tuning (\cite{t5}). This latent memory, however, presents a challenge where a pre-trained language model (PLM) may incorrectly hallucinate a response using its memory rather than adhere to the input document (\cite{neeman2022disentqa}). We explore the performance of retrieve-and-read and end-to-end dense methods in this work, with attention paid to how well a PLM is able to be fine-tuned to minimize downstream hallucinations and promote fact attribution.

%% file: sections/methods.tex
\section{Methods}
\label{methods}

\subsection{Question Types}
A variety of question types are needed to comprehensively test the contextual encoding and generated answer quality of a CQA model.
\begin{itemize}
    \item \textbf{Extractive:} Directly retrieving a span from a given text (e.g. Quoting supporting evidence). This question type further promotes context adherence.
    \item \textbf{Generative:} Generating a response that is potentially novel relative to the input text (e.g. summarizing, yes/no). Generative responses allow for large amount of flexibility and challenges due to the lack of constraint.
    \item \textbf{Unanswerable:} Cases where "unanswerable" should be the generated answer. Requires detection of logical fallacy, insufficient context, and more. These help us test the ability of the proposed models to generate contextually relevant responses without relying on their pre-trained knowledge.
    \item \textbf{Multi-hop}: Questions which involve compositional logical reasoning. These questions necessitate the model to combine information from multiple parts of the context to arrive at the correct answer, further testing the context adherence capabilities of the proposed models.
    \item \textbf{Incomplete Questions}: Cases where not all needed details are mentioned in the question. Here prior or external knowledge is assumed and conditions the response. This is the primary CQA task
\end{itemize}

We opt for a generative solution that allows flexibility with all question types. We seek to work with a dataset that is augmented by each question type and not just incomplete questions in order to comprehensively test the document adherence and reasoning capabilities of our CQA model solution.

\subsection{Context Retrieval}
\label{search_method}
An incomplete question may require knowledge stored within a long context document with the necessary information potentially spread out and related in logically complex ways. This feature allows us to test the ability of the proposed models to process and understand intricate relationships between different pieces of information, which is crucial for context adherence and memory bias reduction. Given the models we are using for our fine-tuned approach, the input context is limited in size and therefore the most relevant facts need to be extracted through a retrieval mechanism. We make use of two retrievers in this work:

\textbf{Oracle Retriever:} To understand the level of performance bottlenecking that occurs due to a subpar retrieval system, we employ an oracle retriever which assumes we have a perfect retrieval mechanism for the downstream model. These downstream models (named tagged with \texttt{goldContext} in later sections) are trained by providing the conditioning evidence, which should be generated in output, as part of the input.

\textbf{Search-based retrieval:} Many fine-tuned LLMs cannot encode a long context document. Instead, we can use an extractive search to only grab the most relevant lines of evidence. This extraction should reduce context noise and facilitate conditioning on the correct evidence. We employ a simple cosine similarity-based search algorithm which is as follows:
\begin{enumerate}
    \item Encode the question (question + scenario) and each line of the document
    \item Rank the lines according to their cosine similarity with the question encoding
\end{enumerate}

\subsection{Fine-tuning for Evidence-based Generation}
Recent GPT LLM performance is exceptional in many tasks including QA. However, hallucinations can occur, especially without fine-tuning as the model defaults to pre-training knowledge which may be stale or biased. We want to create a pipeline with a fine-tuned generative model where the generated response is accurate, and complete and includes justification via extracted evidence. We propose that a fine-tuned approach, especially within a pipeline with other out-of-the-box models, can achieve SOTA performance. We will promote contextual adherence by requiring lines of evidence to be generated as they support the generated answer. Likewise, we seek to support complex questions by targeting multiple answer generations that account for ambiguous cases, while introducing contextual complexity via search-based retrieval. 

\subsection{Evaluation}
For much of our evaluation, we follow the method set buy \cite{cqa} and use Exact Match (EM) and F1 scores. For a given example, EM looks at each token in the generation and target and ranks it positively only if all the tokens match at matching indices. F1 is a function of the precision and recall regarding if generated tokens are part of the target answer.

The CQA dataset evaluates answers based on string matching, which may not be representative in a generative setting. An effective evaluation metric may look at the meaning of the generations rather than specific sequences in order to properly compare the large scope of generative answers. This consideration leads us to use embedding-based evaluation methods. This includes the following:

\begin{itemize}
    \item \textbf{Cosine Similarity:} We make use of a text embeddings model to gather cosine similarities between input question+scenerios and output generations with the implication that high similarity entails high relevance
    \item \textbf{Bert Score:} This is a trained scoring method that sums cosine similarities between two sequence's tokens (\cite{bertScore}). Again the implication is that high similarity entails high relevance
    \item \textbf{Bart Score:} This is a trained scoring method using a seq-seq model that scores how likely one sequence can generate the other \cite{bartScore}. This score directly seeks to rate how likely a given generation can be produced with a given condition.
\end{itemize}

%% file: sections/experiments.tex
\section{Experiments}

\subsection{Dataset}
In this study, we will primarily use the Conditional Question Answering (CQA) dataset to evaluate the performance of the proposed models in addressing memory bias and context adherence challenges (\cite{cqa}). The CQA dataset is specifically designed to test the capabilities of LLMs in handling complex questions with conditional answers, i.e., answers that are only applicable when certain conditions are met. The evidence for these conditions are found within long-context documents which are paired with each question. Questions are formatted to predict the answer as well as examples of evidence. Evidence targets constitute extracted spans rather than novel generations.The original dataset contains 2338 training examples (\texttt{train-full}) and 285 dev examples (\texttt{dev-full}). A total of 652 documents are included with a mean length of 106 lines of evidence per document.

Furthermore, the CQA dataset offers a comprehensive variety of questions, including those that are multi-hop, extractive, yes/no, multiple-choice, and unanswerable. By using the CQA dataset and custom datasets, we aim to provide a thorough evaluation of the proposed models, demonstrating their capabilities in handling complex questions with conditional answers and addressing the challenges associated with memory bias and context adherence in LLMs for conditional question answering. 

\subsubsection{Ambiguous Cases}
 In some examples, multiple answers are presented as valid. To understand the effect of ambiguous cases on model performance, we removed the multiple answer examples from our train and dev datasets to create train-single and dev-single for certain baselines. These ambiguous examples account for 25\% of cases resulting in 1743 and 213 examples for \texttt{train-single} and \texttt{dev-single} respectively.

\subsection{Fine-tune Formatting}
\label{finetuneFormat}
Each example of the CQA dataset has a list of answers, of length 1 or more, with each answer being paired with a list of evidence lines the answer is dependent upon, of length 0 or more. For fine-tuning the input context is formatted as such:

\begin{quote}
\texttt{Instruction: Given the scenario, answer the question and provide evidence for the answer. Scenario is: \{scenario\}. Now the question is: \{question\}}
\end{quote}

Here the scenario is optional dependent on the specific experiment. The target is then as follows:

\begin{quote}
\texttt{Answer is: \{answer1\} My evidence is \{evidence1.1\} || \{evidence1.2\} || … Answer is: \{answer2\} My evidence is \{evidence2.1\} ...}
\end{quote}

Here the evidence and the inclusion of more than one answer are optional and dependent on the experiment.

\subsection{Search-based Context Retrieval}
To retrieve relevant lines of evidence from long-form documents, we implement a basic search algorithm as described in section \ref{search_method}. We explore a set of pre-trained text embeddings models that showed high performance on diverse embedding tasks via \cite{mteb}. To determine the quality of each pre-trained embedding model, we explore how well the cosine-similarity-based ranks overlap with gold evidence lines.

Of the pre-trained models tested, the two top performers are \texttt{all-MiniLM-L6-v2} and \texttt{bge-large-en} (\cite{sentenceTransformers, bge}). In Table \ref{table:embedding} we see comparable performance between the two models with 39.4\% and 39.1\% of gold evidence being in the retrieved top 10 for \texttt{all-MiniLM-L6-v2} and \texttt{bge-large-en} respectively. The performance disparity increases, however, when we examine gold overlap in top1, top3, top5, and top7 rankings. Here \texttt{all-MiniLM-L6-v2} outperforms \texttt{bge-large-en} by about 1\% for each ranking, showing that relatively more of the top10 ranking performance for \texttt{bge-large-en} was weighted at the end of the scale. For this reason, we make use of \texttt{all-MiniLM-L6-v2} for all search-based context retrieval.

\begin{table}
\centering
    \begin{tabular}{c|c|c|c|c|c}
    \hline
    Model & top1 & top3 & top5 & top7 & top10 \\
    \hline
    \texttt{all-MiniLM-L6-v2} & 10 & 20.3 & 26.5 & 31.5 & 39.4 \\
    \texttt{bge-large-en} & 9.1 & 18.9 & 25.2 & 30.1 & 39.1 \\
    \hline
    \end{tabular}
    \caption{Percentage of gold evidence lines present for each embedding-similarity-based ranked list by two pre-trained text-embedding models based on \texttt{train-full}}
    \label{table:embedding}
\end{table}

\subsection{The Models}
Tested model architectures include t5, i.e. \texttt{flan-t5} \texttt{small}, \texttt{base}, \texttt{large}, \texttt{Xl}, and \texttt{xxl}, and \texttt{ul2} (\cite{t5, ul2}). We select T5 due to high performance on this task per previous literature, and ul2 as it is based on T5 and found to outperform T5-XXL on many tasks (\cite{cqa, ul2}). Each model is trained for 10 epochs with the epoch with the best performance over the dev set taken for final evaluation. For \texttt{t5-Xl}, and \texttt{t5-xxl}, Parameter-Efficient-Fine-Tuning (PEFT) was used in order to speed up training times and increase context size (\cite{peft}).

In order to effectively understand the performance of generative models on this dataset, we perform 6 experiments. Each experiment iteratively adds more complexity, allowing insight into bottlenecks and capabilities. Table \ref{table:increment} enumerates each model type and its pertinent input contexts and output targets.

\begin{table}
\centering
    \begin{tabular}{c|c|c|c|c|c|c|c}
    \hline
    \multirow{2}{*}{Model} & \multicolumn{4}{c}{Input} |& \multicolumn{3}{c}{Output} \\\cline{2-8}
    & Question & Scenerio & Gold Evid. & Search Evid. & Answer & Answers & Evid.\\
    \hline
    S1 & \checkmark & & & & \checkmark & & \\
    S2 & \checkmark & \checkmark & & & \checkmark & & \\
    S3 & \checkmark & \checkmark & & & \checkmark & & \checkmark \\
    M1 & \checkmark & \checkmark & & & & \checkmark & \checkmark \\
    M2 & \checkmark & \checkmark & \checkmark & & & \checkmark & \checkmark \\
    M3 & \checkmark & \checkmark & & \checkmark & & \checkmark & \checkmark \\
    \hline
    \end{tabular}
    \caption{Input contexts and Output targets for each experiment. In cases where only one answer is being output, we are making use of the \texttt{train-single} and \texttt{dev-single} datasets for training and evaluation respectively.}
    \label{table:increment}
\end{table}

\subsection{Evaluation}
For EM and F1 evaluation, we make use of the codebase provided by (\cite{cqa}. This scorer takes each example in the dev set and conducts token-wise comparisons before aggregating a score for the dataset as a whole.

For each embedding-based method, we encode the example question+scenerio, as described in \ref{finetuneFormat}. We then encode the predicted generation and provide the two embeddings to each scorer before reporting the mean score for the evaluation dataset. For cosine similarity-based evaluation, we make use of the same model as our search system \texttt{all-MiniLM-L6-v2}.

%% file: sections/results.tex
\section{Results}

We present our results over each experiment (S1-S3 + M1-M3) as well as an investigation of evaluation methods. We follow the results methodology presented in \cite{cqa} and compare our results against the FiD SOTA performance as reported in Table \ref{table:SOTA}. For full results for each of the experiments, see Appendix \ref{completeResults}

\begin{table}
\centering
    \begin{tabular}{c|c|c|c|c}
    \hline
    Category & EM & EM+Cond. & F1 & F1+Cond.\\
    \hline
    \texttt{Yes/No} & 0.64 & 0.48 & 0.64 & 0.48 \\
    \texttt{Extractive} & 0.25 & 0.23 & 0.38 & 0.33 \\
    \texttt{Conditonal} & 0.45 & 0.05 & 0.50 & 0.06 \\
    \texttt{Overall} & 0.44 & 0.35 & 0.51 & 0.41 \\
    \hline
    \end{tabular}
    \caption{SOTA Fusion-in-Decoder results reported by \cite{cqa} and evaluated on the complete test set, \texttt{test-full}}
    \label{table:SOTA}
\end{table}

\subsection{Single-Answer Results}

In the following experiments (S1-S3) the evaluation is conducted on \texttt{dev-single} in order to ascertain performance without ambiguous cases with more than one answer.

\textbf{Experiment S1:} In Table \ref{table:expS1} we see the overall results of experiment S1. In this baseline run, we only consider generations conditioned on the question alone, therefore relying on implicit knowledge these models gather in pre-training and fine-tuning. As expected we see every metric underperforms the FiD SOTA by substantial margins with differences above 10 points in many cases despite the comparatively easier evaluation set (\texttt{dev-single} vs \texttt{test-full}). Larger model size does seem to help performance, though not invariably with \texttt{t5-base} outperforming \texttt{t5-large} in several cases, albeit by only 1-2 points.

\begin{table}
\centering
    \begin{tabular}{c|c|c|c|c}
    \hline
    Model & EM & EM+Cond. & F1 & F1+Cond.\\
    \hline
    \texttt{t5-small} & 0.14 & 0.14 & 0.16 & 0.16 \\
    \texttt{t5-base} & \textbf{0.25} & 0.20 & \textbf{0.32} & \textbf{0.25} \\
    \texttt{t5-large} & \textbf{0.25} & \textbf{0.21} & 0.28 & 0.24 \\
    \hline
    \end{tabular}
    \caption{Experiment S1 overall score results reported on \texttt{dev-single}. [Input: Question, Output: Answer]}
    \label{table:expS1}
\end{table}

\begin{figure}
    \centering
    \includegraphics[width=\linewidth]{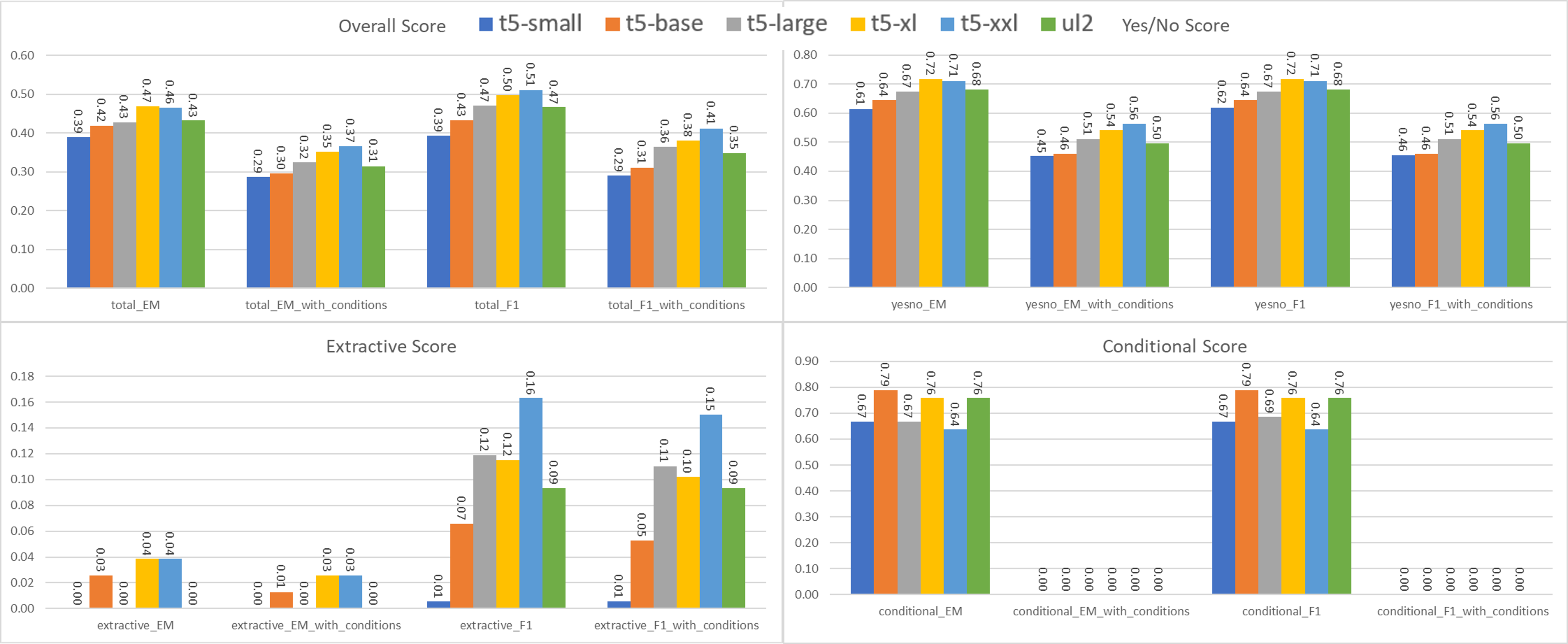}
    \caption{Experiment S2 results reported on \texttt{dev-single}. [Input: Question + Scenario, Output: Answer]}
    \label{fig:expS2}
\end{figure}

\textbf{Experiment S2:} In Figure \ref{fig:expS2} we see the complete results of experiment S2. Here we expand on models created in experiment S1 by including the Scenario alongside the Question in the input context. We again see model size improves performance, with \texttt{t5-xl} and \texttt{t5-xxl} performing best. Meanwhile, we find the \texttt{UL2} model to under-perform against T5 except in Conditional Score cases (bottom right in Figure \ref{fig:expS2}). Furthermore, we see the best Yes/No performance is on track to beat SOTA with Conditional Score also performing well in some cases, albeit these results exclude ambiguous cases which heavily weight Conditional Score and can affect the required reasoning complexity for generating yes/no responses. We find the overall score still has room for improvement even in this easier task, while Extractive Score is far behind SOTA with over 10 point differences in both metrics.

\textbf{Experiment S3:} In Table \ref{table:expS3} we see the overall results of experiment S3. Here we expand on experiment S2 by additionally requiring the model generate lines of evidence to support their answer when appropriate. The trends shown here generally follow those presented with Experiment S2, with relatively higher Yes/No Score and some aspects of Conditional Score but a large gap on Extractive Score and Overall Score. These results, however, show a relative improvement of a few points in most metrics compared to experiment S2. Furthermore, we see that UL2 has trouble generating conditional evidence, marked by the 0.0 scores on \texttt{+Cond} cases. In practice we see this is because the UL2 model simply does not generate these output conditions despite a sufficient generation window and an identical fine-tuning target.

\begin{table}
\centering
    \begin{tabular}{c|c|c|c|c}
    \hline
    Model & EM & EM+Cond. & F1 & F1+Cond.\\
    \hline
    \texttt{t5-xl} & \textbf{0.50} & \textbf{0.39} & \textbf{0.54} & \textbf{0.43} \\
    \texttt{t5-xxl} & 0.49 & \textbf{0.39} & 0.52 & 0.42 \\
    \texttt{ul2} & 0.45 & 0.00 & 0.47 & 0.00 \\
    \hline
    \end{tabular}
    \caption{Experiment S3 overall score results reported on \texttt{dev-single}. [Input: Question + Scenario, Output: Answer + Evidence]}
    \label{table:expS3}
\end{table}

\textbf{Single Answer Findings:} In these experiments (S1-S3) where we train using \texttt{train-single} and evaluate using \texttt{dev-single}, we find overall poor performance on extractive questions and a subset of conditional questions. Poor extractive and conditional performance is expected as the models do not see valid lines of evidence within their input context, meaning getting exact quotes or correct lines of evidence would be near-impossible to generate. Furthermore, we find \texttt{t5-xxl} and \texttt{t5-xl} often outperforms UL2, especially in case of evidence generation (conditional answers), due to UL2 not generating evidence lines whatsoever. We find including the scenerio in the input and evidence in the output improved model performance in nearly all metrics. Performance overall is approaching SOTA while Yes/No performance is beginning to exceed SOTA performance, albeit not directly comparable due to the filtered evaluation set, \texttt{dev-single}, which removes ambiguous questions.

\subsection{Multi-Answer Results}

For the following experiments (M1-M3), we evaluate each model using \texttt{dev-full} and train most models using \texttt{train-full}, therefore including ambiguous cases where multiple answers may be valid.

\begin{figure}
    \centering
    \includegraphics[width=\linewidth]{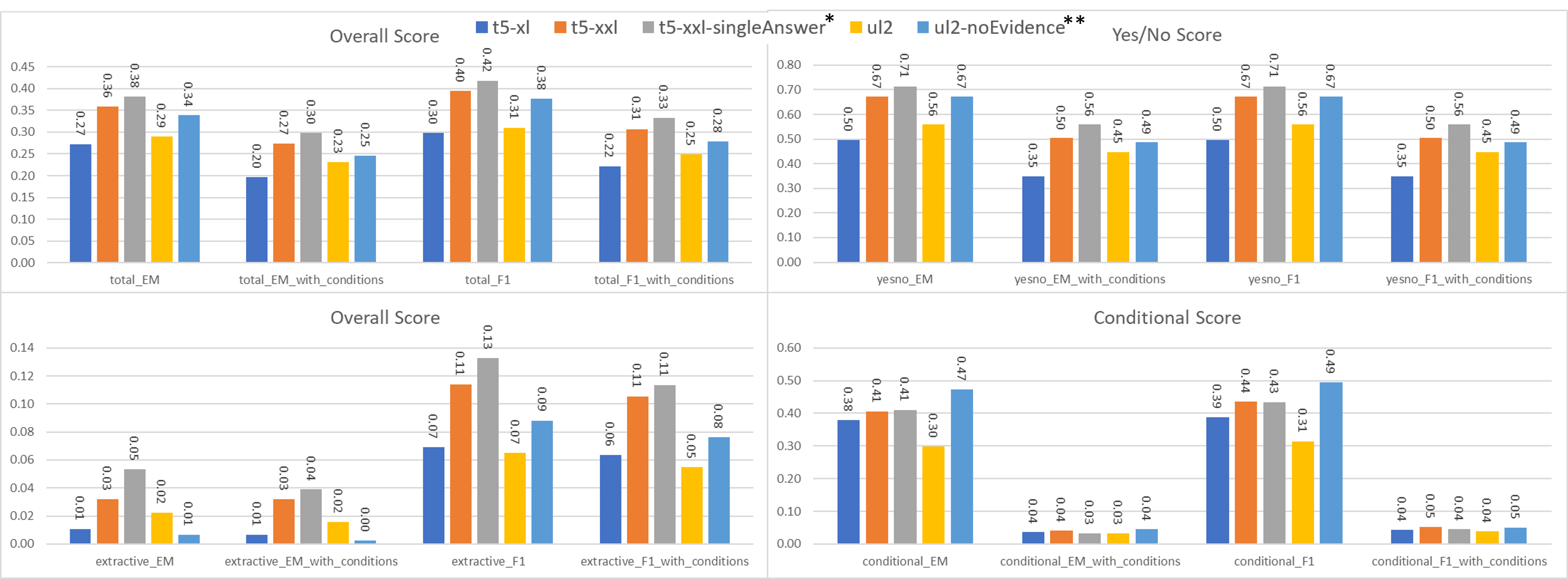}
    \caption{Experiment M1 results reported on \texttt{dev-full}. \texttt{t5-xxl-singleAnswer} is trained to output only a single answer while \texttt{ul2-noEvidence} is trained to output answers without evidences. [Input: Question + Scenario, Output: Answers* + Evidence**]}
    \label{fig:expM1}
\end{figure}

\textbf{Experiment M1:} In Figure \ref{fig:expM1} we see the complete results of experiment M1. Here we expand on experiment S3 by training models to output multiple answers, and their relevant evidence lines, rather than just 1 answer. Surprisingly, \texttt{t5-xxl-singleAnswer}, which only ever attempts to output one answer, often performs best despite evaluation now being done over \texttt{dev-full}. This may be a function of the dataset or evaluation metric not penalizing single-answer generation appropriately. Compared to the SOTA, \texttt{t5-xxl-singleAnswer} is lower by about 6 points in the overall metric. \texttt{t5-xxl-singleAnswer} Conditional Score matches or slightly underperforms SOTA, while \texttt{ul2-noEvidence} seems to beat SOTA in some cases. \texttt{t5-xxl-singleAnswer} Yes/No performance outperforms SOTA while Extractive performance in all models continues to far underperform SOTA.

We see \texttt{t5-xxl} also performs well while \texttt{ul2} underperforms against \texttt{t5}. This trend holds except when \texttt{ul2} is not trained to output evidence, in which case the performance starts to match or exceed in some cases. Since no evidence is being extracted, however, these model will have a hard cap on their maximum performance as they can only be 100\% correct in cases with no conditional evidence supporting a given answer.

\begin{figure}
    \centering
    \includegraphics[width=\linewidth]{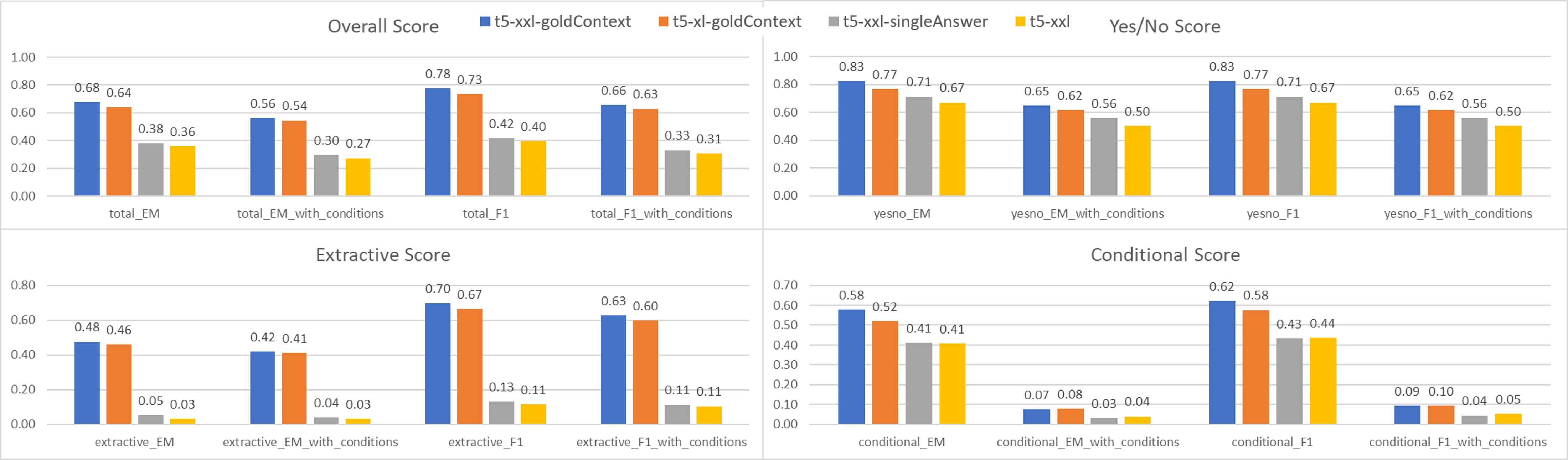}
    \caption{Experiment M2 results reported on \texttt{dev-full}. [Input: Question + Scenario + Gold-Evidence, Output: Answers + Evidence]}
    \label{fig:expM2}
\end{figure}

\textbf{Experiment M2:} In Figure \ref{fig:expM2} we see the complete results of experiment M2. Here we expand on experiment M1 by including lines of evidence within the model's input context during training. Namely, we assume an Oracle retriever which allows us to have the gold-labelled evidences (i.e. target evidences) directly added to the input context. As expected, providing gold context substantially increases performance of the top models, well above the SOTA. Performance gain on Yes/No is minimal while extractive gains are transformative. Even so, Many of the scores, while improving above SOTA, are still low overall with scores ranging from 0.5-0.8 depending on the score metric. This shows both that evidence retrieval is a major bottleneck in the performance of our QA models, and furthermore, that even with an Oracle retriever, performance can continue to improve in other areas such as reasoning.

\textbf{Experiment M3:} We present overall results for experiment M3 in Table \ref{table:expM3}. Here we modify experiment M2 by having the included input evidence retrieved via an embedding-based search algorithm rather than a presumed Oracle retriever. We show performance when trained on different quantities of retrieved evidences, based on top-k ranking. We see \texttt{t5-xl-topK3}, \texttt{t5-xxl-topk7}, \texttt{t5-xxl-topK3} are the best performers, with \texttt{t5-xl-topK5} having the highest scores overall. Given the top performers range in model size and evidence length, it is not clear how to determine the best K to select which balances encoding size and relevant information. This balance is especially relevant as given the results from Table \ref{table:embedding} which show many of the top 10 results for embedding-based retrieval do not actually overlap with the target evidence, indicating that with higher K values, more mixed irrelevant evidences will be provided, potentially confusing the context signal for the model.

\begin{table}
\centering
    \begin{tabular}{c|c|c|c|c}
    \hline
    Model & EM & EM+Cond. & F1 & F1+Cond.\\
    \hline
    \texttt{t5-xl-topK3} & \textbf{0.36} & \textbf{0.27} & \textbf{0.39} & \textbf{0.31} \\
    \texttt{t5-xl-topK5} & 0.12 & 0.10 & 0.14 & 0.12 \\
    \texttt{t5-xl-topK7} & 0.13 & 0.11 & 0.16 & 0.13 \\
    \texttt{t5-xl-topK10} & 0.20 & 0.15 & 0.22 & 0.17 \\
    \texttt{t5-xxl-topK3} & 0.29 & 0.23 & 0.32 & 0.25 \\
    \texttt{t5-xxl-topK5} & 0.15 & 0.12 & 0.15 & 0.12 \\
    \texttt{t5-xxl-topK7} & 0.33 & \textbf{0.27} & 0.35 & 0.28 \\
    \hline
    \end{tabular}
    \caption{Experiment M3 overall score results reported on \texttt{dev-full}. [Input: Question + Scenario + Search-Evidence, Output: Answers + Evidence]}
    \label{table:expM3}
\end{table}

\textbf{Best Performers:} Table \ref{table:expBestPerf} shows the results of the best performing models. According to the evaluation metrics of Exact Match (EM) and F1, we see \texttt{t5-xxl-singleAnswer} performs best despite not being trained to output several answers in ambiguous cases. The maintained top performance of \texttt{t5-xxl-singleAnswer} could indicate more complex generation targets can actually hurt performance even though having multiple answer generations is required to appropriately cover each question type in the CQA dataset. Furthermore, we see the inclusion of search-based evidence retrieval seems to reduce performance by about 1 point compared to simply \texttt{t5-xxl}. This finding could indicate that the inclusion of noisy evidence may confuse the encoded signal for the model, causing worse subsequent answer generations compared to simply relying on Question/Scenario information and falling back to parametric knowledge via fine-tuning and pre-training steps.

\begin{table}
\centering
    \begin{tabular}{c|c|c|c|c}
    \hline
    Model & EM & EM+Cond. & F1 & F1+Cond.\\
    \hline
    \texttt{(S3) t5-xxl-singleAnswer} & \textbf{0.38} & \textbf{0.30} & \textbf{0.42} & \textbf{0.33} \\
    \texttt{(M1) t5-xxl} & 0.36 & 0.27 & 0.40 & 0.31 \\
    \texttt{(M3) t5-xl-topK3} & 0.36 & 0.27 & 0.39 & 0.31 \\
    \texttt{(M3) t5-xxl-topK7} & 0.33 & 0.27 & 0.35 & 0.28 \\
    \hline
    \end{tabular}
    \caption{Best performers overall score results reported on \texttt{dev-full}. The version of each model is marked in parentheses. [Input: Question + Scenario + (Gold/Search/No)-Evidence, Output: Answers + Evidence]}
    \label{table:expBestPerf}
\end{table}

\subsection{Other Evaluation Methods}
The generative nature of our QA systems, alongside the peculiarly high performance of the \texttt{t5-xxl-singleAnswer} model on Exact-Match and F1 indicates that a semantic evaluation technique may allow further insights into model performance that token-wise matching evaluation methods miss out on. Table \ref{table:otherEval}, which presents the results of our best performing models when evaluated using embedding-based techniques, affirms this hypothesis. We see in this case that while \texttt{t5-xxl-single} is still a good performer, achieving the second highest score with each metric, \texttt{t5-xxl-topK7} ends up being the top-ranked performer with each metric. 

These results indicate that a comprehensive evaluation is needed to show the whole story as to the best-performing model, rather than just token-matching or embedding-based methods alone. Even so, the performance discrepancy between each model is fairly small, with only a few points of separation for each metric. Instead we see \texttt{t5-xxl-goldContext} continues to have a substantial lead in each metric compared to non-oracle models, indicating again that retrieval is a major bottleneck in the CQA task.

\begin{table}
\centering
    \begin{tabular}{c|c|c|c}
    \hline
    Model & Cosine Similarity & Bert Score & Bart Score\\
    \hline
    \texttt{(M2) t5-xxl-goldContext} & 0.80 & 0.95 & -2.15 \\
    \texttt{(M3) t5-xxl-topK7} & \textbf{0.68} & \textbf{0.93} & \textbf{-2.99}\\
    \texttt{(S3) t5-xxl-singleAnswer} & 0.62 & 0.92 & -3.22 \\
    \texttt{(M3) t5-xl-topK3} & 0.62 & 0.92 & -3.25 \\
    \texttt{(M1) t5-xxl} & 0.62 & 0.92 & -3.26 \\
    \hline
    \end{tabular}
    \caption{Best performers embedding-based evaluation results reported on \texttt{dev-full}. GoldContext (M2) model provided as reference showing oracle performance. The version of each model is marked in parentheses. [Input: Question + Scenario + (Gold/Search/No)-Evidence, Output: Answers + Evidence]}
    \label{table:otherEval}
\end{table}

%% file: sections/analysis.tex
\section{Analysis}

We find that SOTA on Yes/No questions can be beaten using fine-tuned generative LLMs without adjustments to model architecture, with \texttt{t5-xxl-singleAnswer} outperforming SOTA by 7-8 points in terms of EM and F1. This is a reasonable finding as Yes/No responses are the most directly generative component, requiring reasoning over input context and output of a single novel response. With regards to other question types, we see our fine-tuned models may come close to SOTA, although extractive responses remain difficult for generative models.

Additionally, we find selecting the best evidence to condition the generated response on is a critical challenge. In our oracle-retriever experiments, we see a large gap indicated a bottleneck due to evidence retrieval alone. Our implement search-retrieval methods, however, did little to close this gap, indicating more sophisticated solutions are likely necessary.

Our embedding-based evaluation identified that evaluation methods can skew perspective on performance, with token-wise and semantic evaluations disagreeing on what the top performing model is. It is difficult to ascertain which automatic metric is the 'most correct' as various functional properties may be desired. An answer that is semantically correct may be sufficient in some cases whereas other cases a user may require an explicit quote of evidence to verify model generation. It is likely that a more comprehensive evaluation suite will be necessary.

Finally, we see that both for the Oracle-evidence and SOTA models, there is still much room for improvement. Without human baselines to compare against, it is difficult to see how much of the performance gap left is due to faulty dataset examples or truly model performance deficiencies. Regardless, the performance gap and the complex nature of the dataset alone continue to indicate the challenging nature of the task.

%% file: sections/conclusion.tex
\section{Conclusion and Future Work}
\label{conclusion}

In this paper we explore the current state of Large Language Models (LLMs) in conditional question-answering tasks, particularly on the CQA dataset and using generative LLMs like T5 and UL2.

We find beating SOTA is possible with a simple fine-tuned LLM despite not effectively encoding all the needed input context. However, extractive answers remain challenging for generative models. We highlight the critical challenge of selecting the best evidence for conditioning-generated responses and suggest that more sophisticated methods may be required.

Furthermore, we underscore the impact of evaluation methods on performance perspectives, as different metrics can disagree on the top-performing model. We argue the need for a more comprehensive evaluation suite and note how without human baselines for comparison, it's challenging to determine the extent of performance gaps, making it clear that the task remains complex and challenging.

In future work, we will explore solutions that manipulate the training task to be focused on extraction of evidence at first, followed by the generation of answers at a later fine-tuning step, ideally forcing the model to better retrieve and pass-forward evidence. Furthermore, future work can make use of prompt-based methods such as self-correction or few-shot answer generation.

%% file: sections/appendix.tex
\section{Appendix}

\subsection{Complete Results}
\label{completeResults}
\begin{figure}
    \centering
    \includegraphics[width=\linewidth]{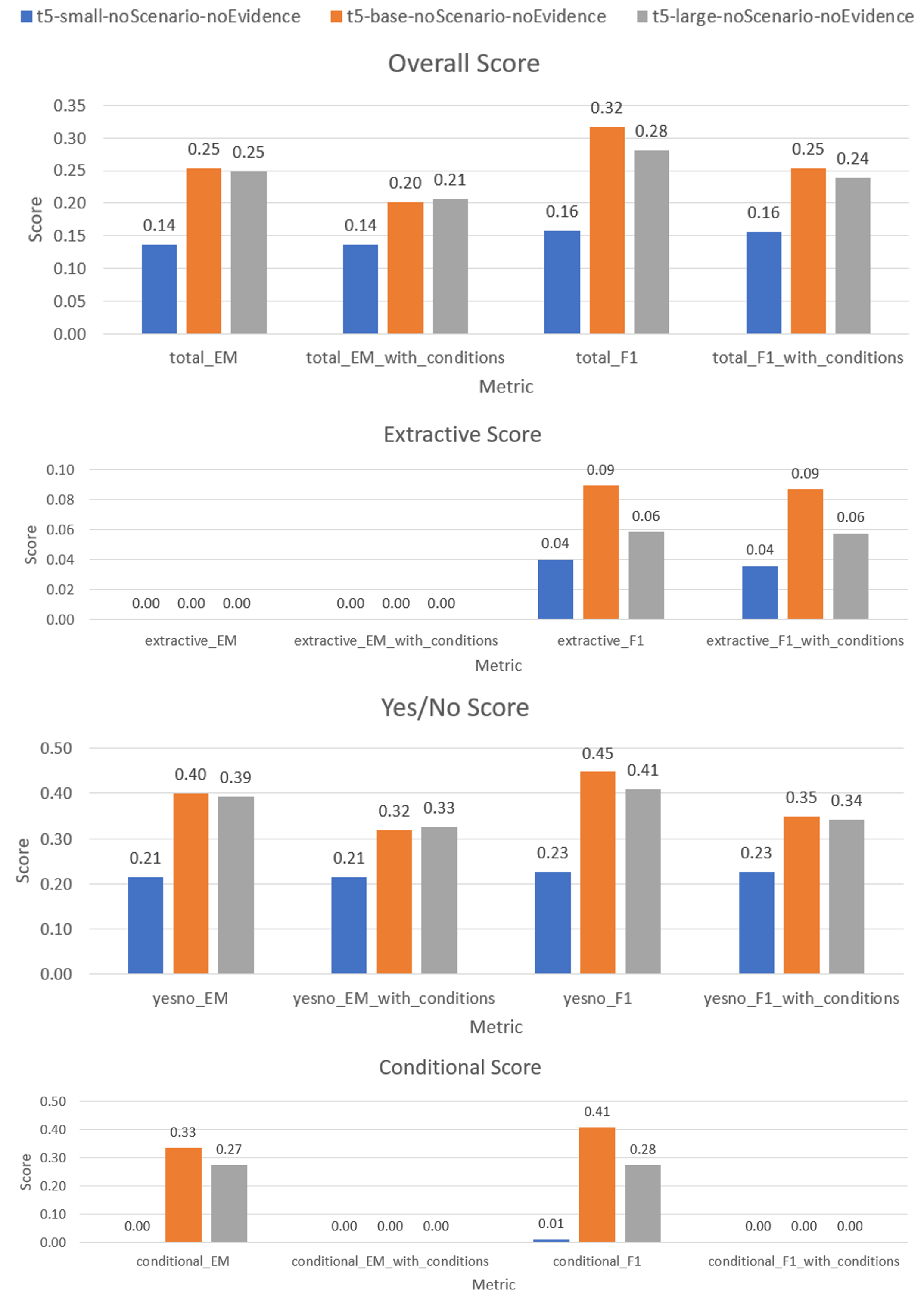}
    \caption{Experiment S1 complete results reported on \texttt{dev-single}. [Input: Question, Output: Answer]}
    \label{fig:S1_Vertical}
\end{figure}

\begin{figure}
    \centering
    \includegraphics[width=\linewidth]{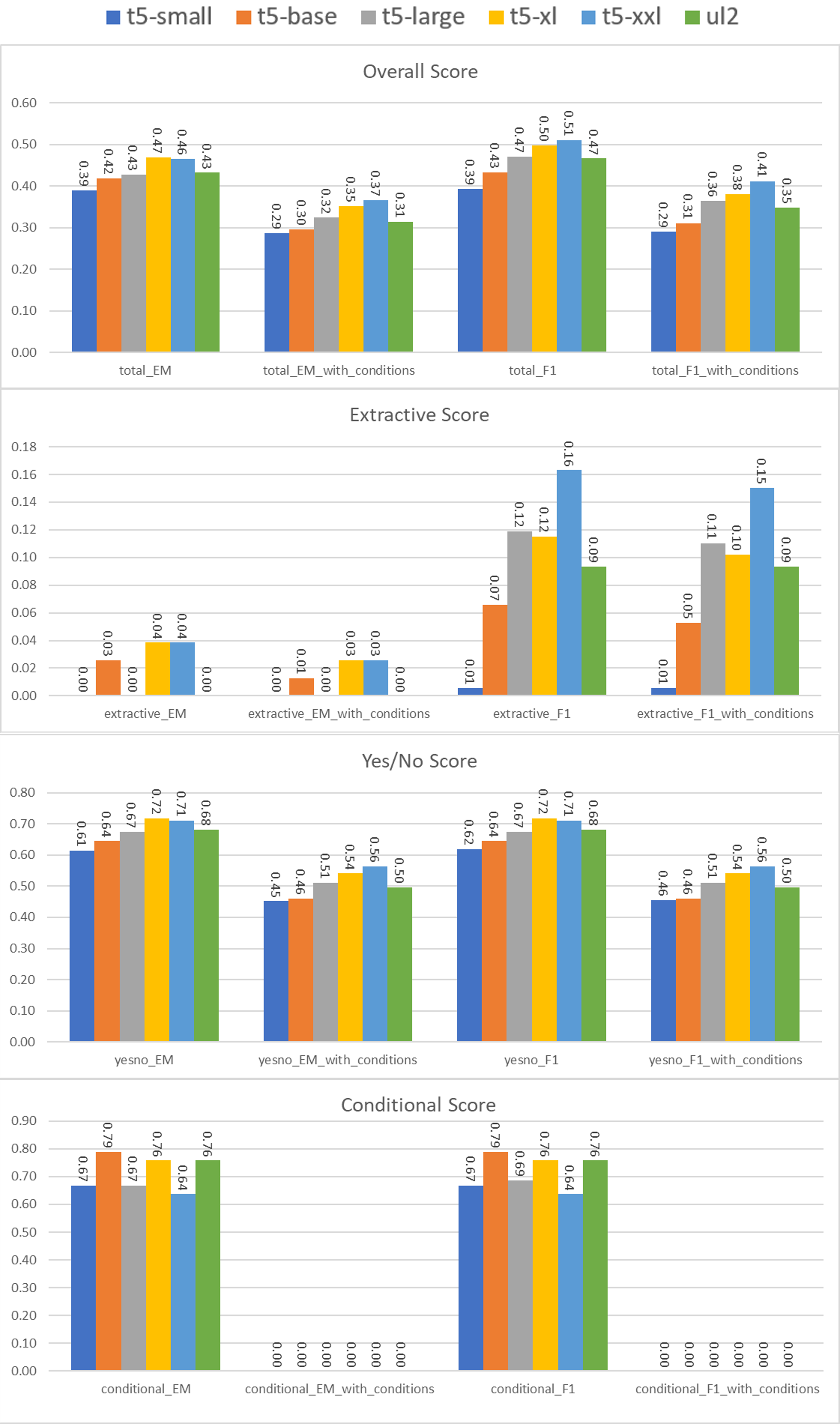}
    \caption{Experiment S2 complete results reported on \texttt{dev-single}. [Input: Question + Scenario, Output: Answer]}
    \label{fig:S2_Vertical}
\end{figure}

\begin{figure}
    \centering
    \includegraphics[width=\linewidth]{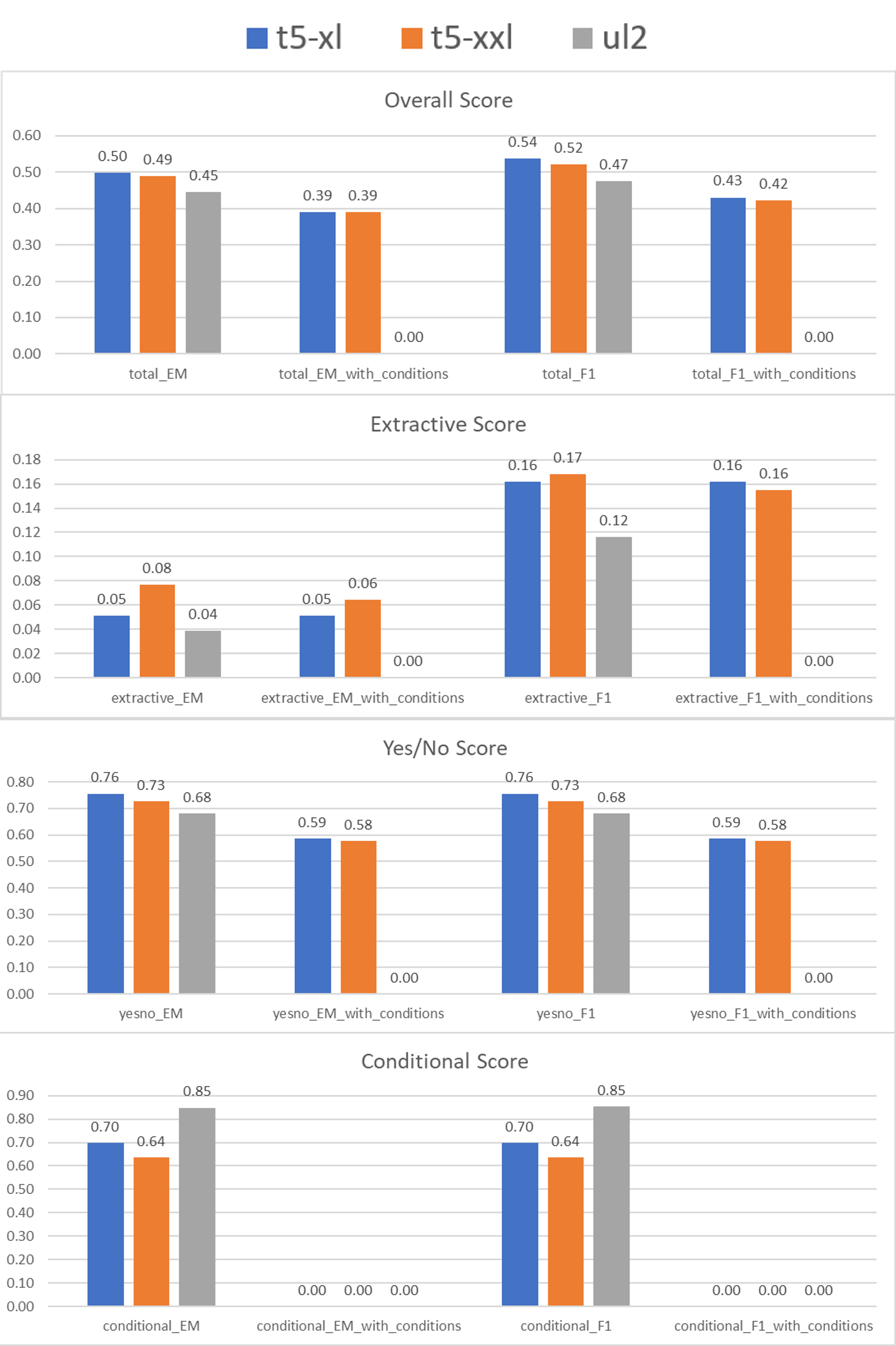}
    \caption{Experiment S3 complete results reported on \texttt{dev-single}. [Input: Question + Scenario, Output: Answer + Evidence]}
    \label{fig:S3_Vertical}
\end{figure}

\begin{figure}
    \centering
    \includegraphics[width=\linewidth]{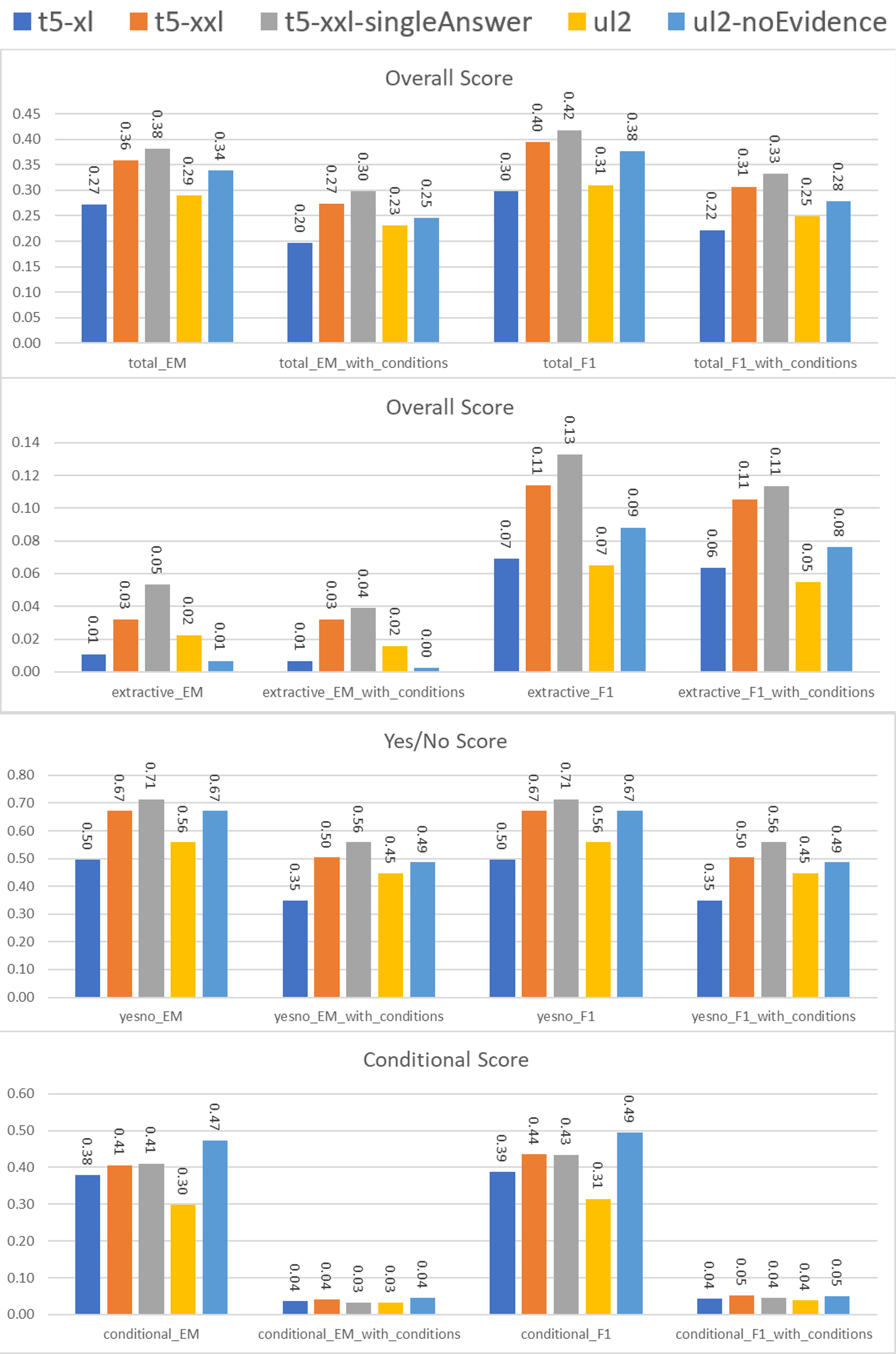}
    \caption{Experiment M1 complete results reported on \texttt{dev-full}. \texttt{t5-xxl-singleAnswer} is trained to output only a single answer while \texttt{ul2-noEvidence} is trained to output answers without evidences. [Input: Question + Scenario, Output: Answers* + Evidence**]}
    \label{fig:M1_Vertical}
\end{figure}

\begin{figure}
    \centering
    \includegraphics[width=\linewidth]{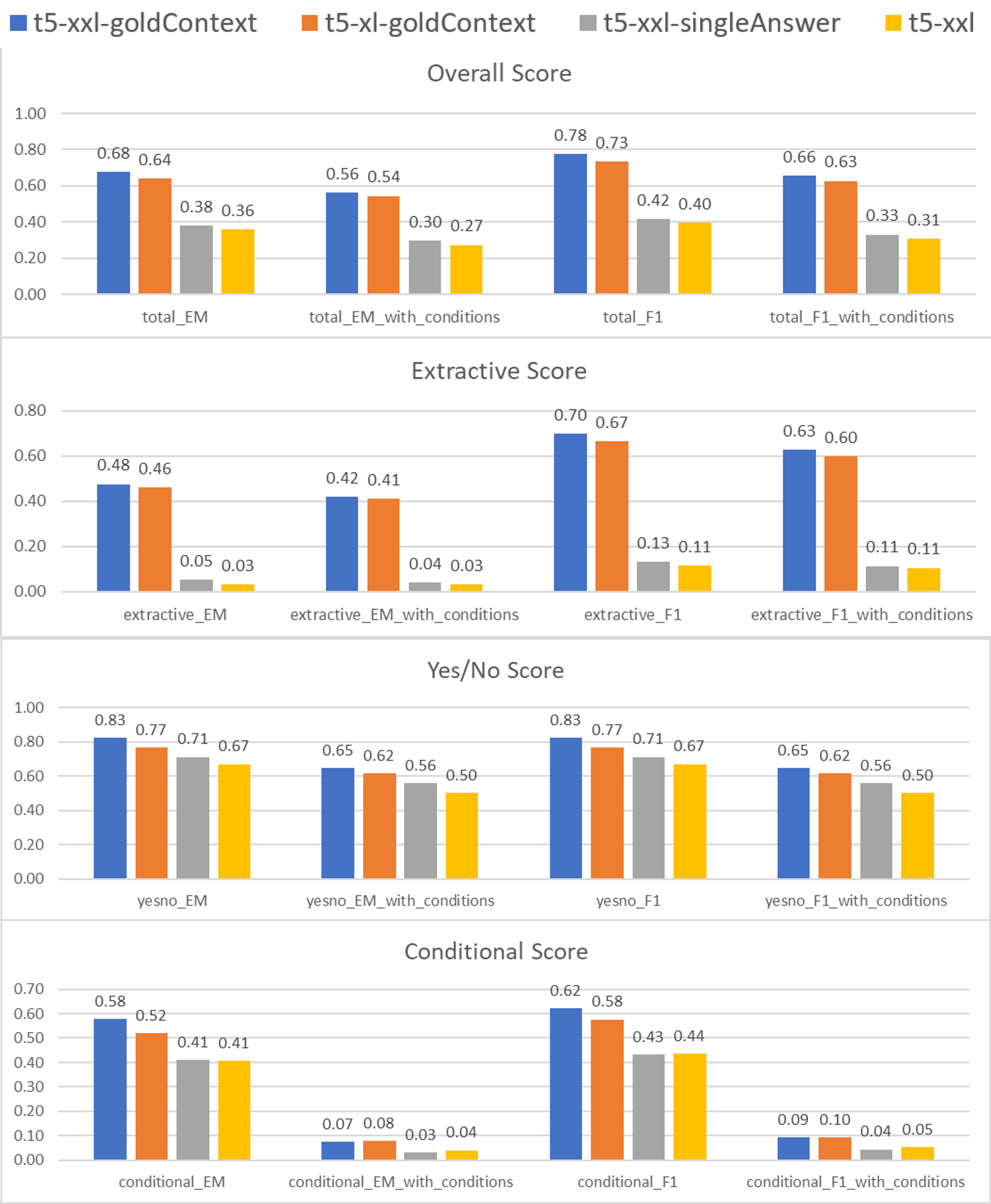}
    \caption{Experiment M2 complete results reported on \texttt{dev-full}. [Input: Question + Scenario + Gold-Evidence, Output: Answers + Evidence]}
    \label{fig:M2_Vertical}
\end{figure}

\begin{figure}
    \centering
    \includegraphics[width=\linewidth]{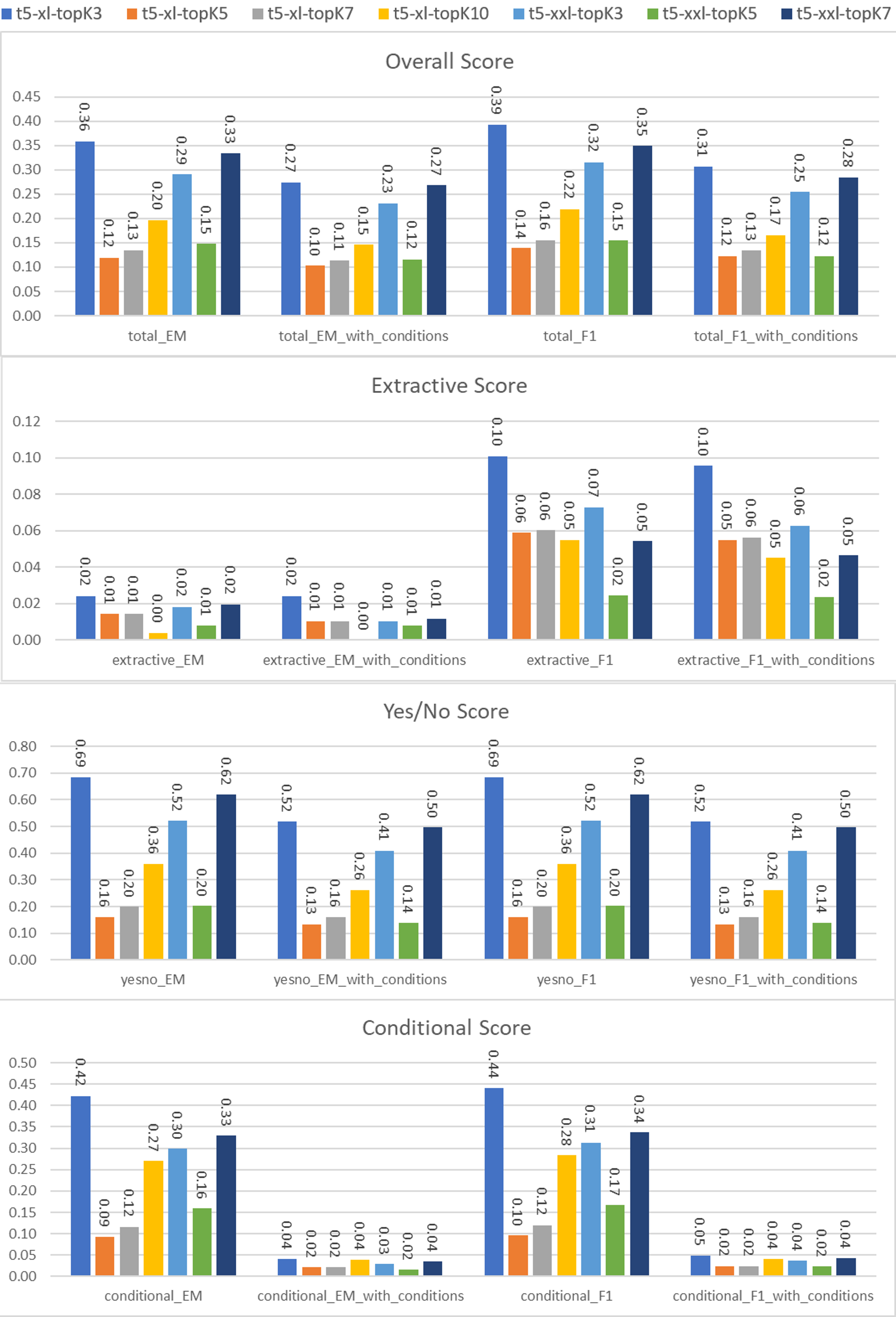}
    \caption{Experiment M2 results reported on \texttt{dev-full}. [Input: Question + Scenario + Gold-Evidence, Output: Answers + Evidence]}
    \label{fig:M3_Vertical}
\end{figure}

\begin{figure}
    \centering
    \includegraphics[width=\linewidth]{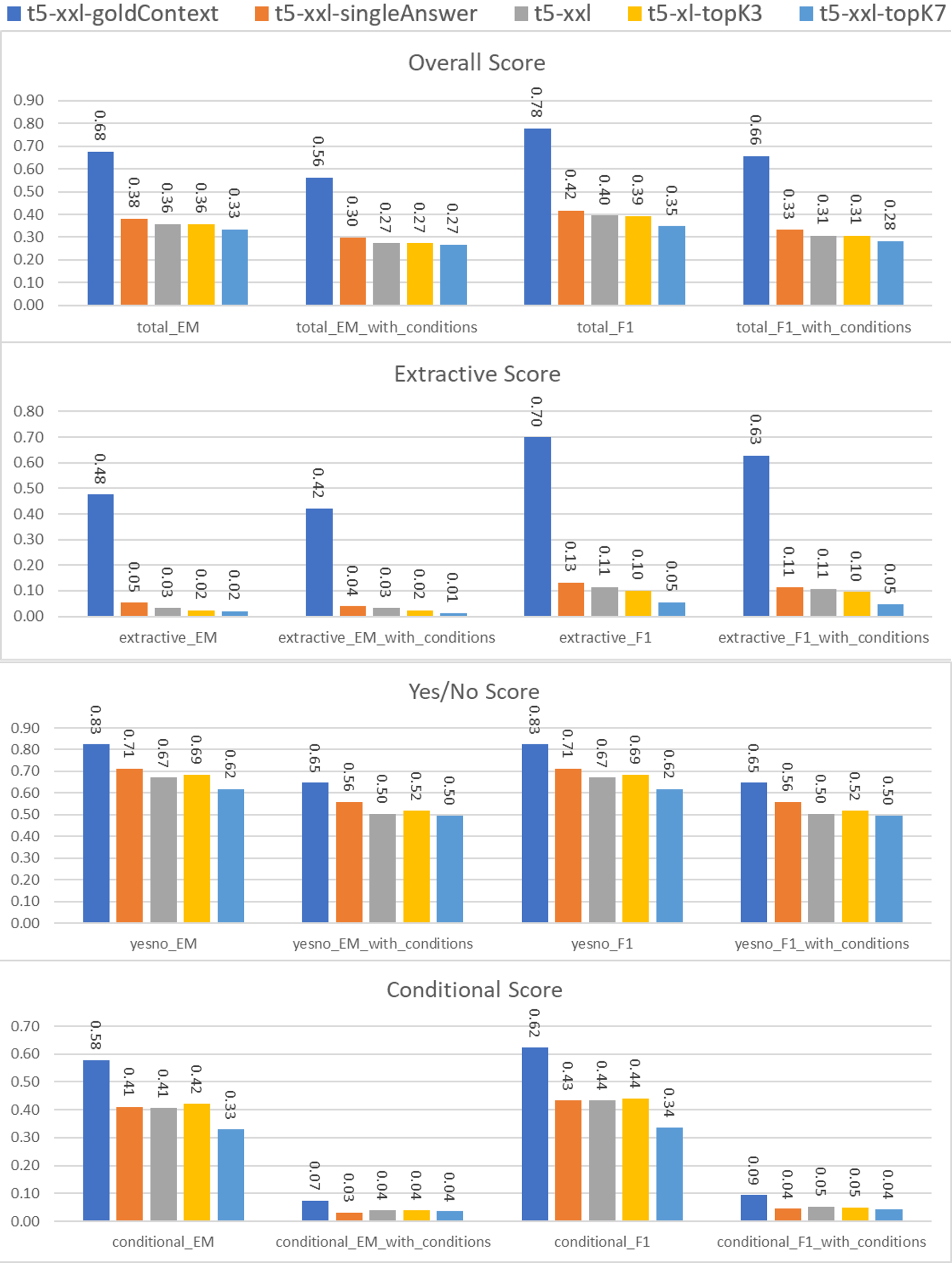}
    \caption{Best performers complete results reported on \texttt{dev-full}. The version of each model is marked in parentheses. [Input: Question + Scenario + (Gold/Search/No)-Evidence, Output: Answers + Evidence]}
    \label{fig:BestPerf_Vertical}
\end{figure}